\documentclass{article} 
\usepackage{iclr2027_conference,times}

\usepackage{amsmath,amsfonts,bm}

\def\eqref#1{equation~\ref{#1}}

\def\1{\bm{1}}

\DeclareMathAlphabet{\mathsfit}{\encodingdefault}{\sfdefault}{m}{sl}
\SetMathAlphabet{\mathsfit}{bold}{\encodingdefault}{\sfdefault}{bx}{n}

\usepackage{hyperref}
\usepackage{url}
\usepackage{xcolor}
\usepackage{algorithm}
\usepackage{algpseudocode}
\usepackage{comment}

\usepackage{booktabs}
\usepackage{graphicx}

\usepackage{tabularx}
\definecolor{wthi}{HTML}{D7301F}
\newcommand{\wt}[2]{{\setlength{\fboxsep}{1pt}%
\colorbox{wthi!#1!white}{\strut\small\texttt{#2}}}}

\newcommand{\std}[1]{\textcolor{gray}{\scriptsize $\pm$ #1}}

\title{GAW-PO: Preference Optimization with Gradient-Aligned Token Weights}

\author{
Andreea Dutulescu$^{1}$,
Stefan Ruseti$^{1}$,
Mihai Masala$^{1}$,
Traian Rebedea$^{1,2}$,
Mihai Dascalu$^{1}$
\\
$^{1}$National University of Science and Technology POLITEHNICA Bucharest \\
$^{2}$NVIDIA
}

\iclrfinalcopy 
\begin{document}

\maketitle

\begin{abstract}
Most preference optimization methods, such as Direct Preference Optimization (DPO), apply preference supervision at the response level, although autoregressive language models are optimized token by token. As a result, all tokens in a rejected response contribute to the negative training signal, including tokens that may encode behavior that is useful for the preferred response. We introduce GAW-PO, a gradient-aligned token reweighting method for DPO that estimates, for each rejected token, whether penalizing it would interfere with the preferred update directions. Tokens whose gradients are strongly aligned with the preferred behavior receive a weaker negative contribution, while conflicting tokens retain a stronger penalty. Our method achieves the highest average performance among the evaluated preference-optimization methods, improving by 0.97 points over standard DPO and 0.65 points over the strongest competing baseline across 11 benchmarks spanning mathematics, reasoning, coding, and question answering. We further show that gradient-aligned weighting is substantially more robust to aggressive preference optimization: as the DPO regularization parameter $\beta$ decreases, standard DPO degrades sharply, whereas GAW-PO continues to improve. These results suggest that accounting for the interaction between rejected-token updates and preferred behavior provides an effective form of token-level credit assignment for preference optimization.
\end{abstract}

\section{Introduction}

Preference optimization has become a standard approach for aligning Large Language Models (LLMs) with human preferences. Among these methods, Direct Preference Optimization (DPO)~\citep{rafailov2023dpo} is widely used due to its simplicity and effectiveness. Given a prompt, a chosen (preferred) response, and a rejected response, DPO trains the model to increase the relative preference for the chosen completion over the rejected one with respect to a reference policy. However, the supervision available to DPO is defined at the response level, while autoregressive language models operate at the token level. As a consequence, all tokens in a rejected response contribute to the optimization signal, even though only a small subset of them may be responsible for the response being considered of lesser quality.

\begin{figure*}[t]
    \centering
    \includegraphics[width=\textwidth]{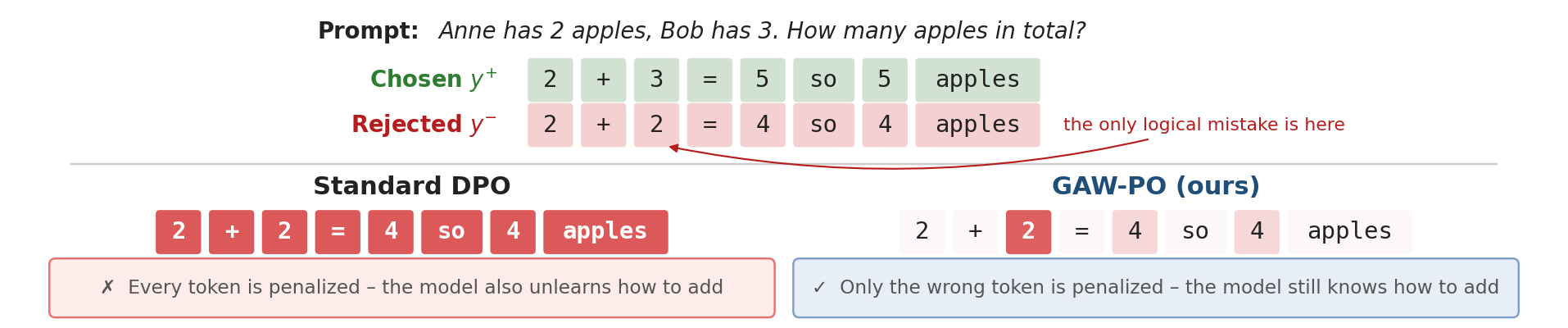}
    \caption{
    Standard DPO penalizes every token of the rejected response, including the arithmetic; our method concentrates the penalty on the token that conflicts with the preferred response.
    }
    \label{fig:teaser}
\end{figure*}

This mismatch can lead to undesirable or at least inefficient updates during preference optimization. A rejected completion is rarely incorrect in its entirety: it may contain valid facts, correct intermediate calculations, appropriate formatting, or linguistic patterns that are also useful for generating the preferred response. Nevertheless, standard DPO does not explicitly distinguish these locally correct components from the tokens that actually determine the preference. Reducing the likelihood of the rejected response can therefore also suppress behaviors that should ideally be preserved. Figure~\ref{fig:teaser} illustrates a simple example of this effect. More generally, the DPO objective constrains the relative preference between the chosen and rejected completions, but does not require the absolute likelihood of the chosen response to increase. Indeed, several works have observed that the likelihood of preferred responses can decrease during DPO training even while the preference objective improves~\citep{pal2024smaug,pang2024irpo}. This behavior suggests that the optimization dynamics of preferred and rejected responses can be strongly coupled~\citep{razin2025unintentional}.

This limitation can be addressed by introducing finer-grained supervision at the token level. Recent methods have proposed token-level preference objectives~\citep{zeng2024tdpo} or weighting schemes~\citep{liu2025tis,yang2026tidpo,li2025aao} that avoid treating every token in a response as equally informative. These methods primarily aim to estimate how important or informative an individual token is. Existing methods alone, however, do not fully characterize the consequence of penalizing that token during optimization.

Conversely, we consider the effect that suppressing a rejected token would have on the model's ability to represent the preferred response. Our starting point is the observation that training samples interact through their parameter updates. This idea has a long history in machine learning, most prominently in work on influence functions~\citep{koh2017influencefc} and gradient-based influence estimation~\citep{pruthi2020estimating}. In these approaches, the relationship between training examples can be characterized by comparing the parameter updates they induce. Examples with similar gradient directions tend to modify the model in compatible ways, while opposing gradients indicate conflicting training signals. Related ideas have also been applied to LLMs, for example for selecting instruction-tuning or reasoning data based on gradient similarity~\citep{xia2024less,jung2026prismatic}.

We apply this perspective to token-level preference optimization. Rather than assigning a weight to a rejected token based only on its probability, loss, or gradient magnitude, we investigate how the update induced by penalizing that token interacts with the update supported by the preferred response. If a rejected token induces a parameter update that is strongly aligned with the one associated with the preferred completion, suppressing that token is likely to interfere with behavior that is also useful for modeling the preferred response. Such a token should therefore receive a weaker negative signal. Conversely, if the token induces an update that is dissimilar to, or conflicts with, the preferred-response update, penalizing it is less likely to damage the behavior encouraged by the chosen completion and can therefore receive a stronger negative signal.
Based on this principle, we introduce GAW-PO, a gradient-aligned token reweighting method for DPO that explicitly accounts for interference between rejected-token updates and preferred update directions. 

Our main contributions are: \textbf{(1)} We propose a gradient-aligned token weighting objective for DPO that selectively downweights rejected tokens whose penalization would interfere with behavior supported by preferred responses. \textbf{(2)} We show that our formulation improves average downstream performance over standard DPO and competing methods, while remaining stable under substantially more aggressive optimization regimes in which standard DPO degrades sharply. We open-source our implementation, gradient-weight computation pipeline, training recipes, filtered preference data, and evaluation setup to facilitate reproducibility and further study: \url{https://anonymous.4open.science/r/Weighted-DPO-Supplementary-8441/}

\section{Related Work}
\textbf{Preference optimization.}
DPO~\citep{rafailov2023dpo} replaced the reward model and reinforcement learning stage of RLHF~\citep{christiano2017deep,ouyang2022instructgpt} with a single classification loss on preference pairs. Most of its variants keep the entire response as the unit of optimization: IPO~\citep{azar2024ipo} changes the loss to limit overfitting, KTO~\citep{ethayarajh2024kto} learns from unpaired feedback, and SimPO~\citep{meng2024simpo} replaces the reference model with a length-normalized reward. A known side effect of this response-level objective is that the likelihood of the chosen response can decrease during training, in particular when the two responses differ in only a few tokens~\citep{pal2024smaug,pang2024irpo,razin2025unintentional}. The previously proposed solutions also act on whole responses, either by adding a likelihood term on the chosen response~\citep{pal2024smaug,pang2024irpo} or by filtering out pairs with similar representations~\citep{razin2025unintentional}. Our method keeps every pair and changes only how strongly each rejected token is penalized.

\textbf{Token- and step-level credit assignment.}
Finer-grained methods differ mainly in the source of the token-level signal. TDPO~\citep{zeng2024tdpo} rewrites the Bradley--Terry objective in terms of token-level advantages but still weights all tokens equally, while TIS-DPO~\citep{liu2025tis} takes token weights from the probability gap between two contrastive LLMs, which have to be obtained by prompting or trained separately. For reasoning, Step-DPO~\citep{lai2024stepdpo} moves from tokens to steps: it finds the first incorrect step of a wrong solution and contrasts a correct and an incorrect continuation of the same correct prefix. This leaves the shared steps unpenalized, but requires locating the error, manually or using a strong LLM judge, and resampling continuations. Similar to our approach, TI-DPO~\citep{yang2026tidpo} also relies on gradients, but computes them with respect to the input embeddings and weights each token by the gradient norm, which measures how strongly the output depends on the token and not whether penalizing it would work against the chosen response.

AAO~\citep{li2025aao} is motivated by the same problem as our method: DPO pushes the same content up in the chosen response and down in the rejected one. It compares each token with the tokens of the other response using the model's input embeddings, and assigns lower weights to tokens that have a close match. However, input embeddings ignore context, so a rejected token is treated as shared whenever the same token appears anywhere in the chosen response. In Figure~\ref{fig:teaser}, this includes the incorrect operand ``2'', which also appears as the first operand of the chosen response.

\textbf{Gradient interaction between training signals.}
GAW-PO measures the interaction between rejected-token and chosen-response updates directly. Following the use of gradient inner products to relate training examples in influence estimation and data selection~\citep{koh2017influencefc,pruthi2020estimating,xia2024less}, we score each rejected token by the inner product between its parameter gradient and the gradients of chosen responses, from its own pair and averaged over the training dataset. Unlike the TI-DPO and AAO scores, this score is signed and depends on response context, and it is computed once before training without an auxiliary model. The closest work in terms of intent is NTHR~\citep{deng2025nthr}, which reduces the penalty on tokens of incorrect responses in GRPO by approximating their gradient interactions through hidden-state and prediction-error alignment.
Most token-level methods above were evaluated on dialogue, safety or summarization with LLM judges, or on a single domain, while we evaluate on 11 benchmarks across mathematics, reasoning, code, and question answering. 

\section{Method}
\label{sec:method}

\subsection{Preliminaries}

We consider a preference dataset $\mathcal{D}={(x,y^{+},y^{-})}$, where $x$ denotes the prompt, $y^{+}$ the chosen response, and $y^{-}$ the rejected response. Direct Preference Optimization (DPO)~\citep{rafailov2023dpo} optimizes a policy $\pi_{\theta}$ relative to a fixed reference policy $\pi_{\mathrm{ref}}$ according to the following loss:

\begin{equation}
\label{eq:dpo_formula}
\mathcal{L}_{\mathrm{DPO}}
=
-\mathbb{E}_{(x,y^{+},y^{-})\sim\mathcal{D}}
\left[
\log \sigma
\left(
\beta
\left[
\log\frac{\pi_{\theta}(y^{+}\mid x)}
{\pi_{\mathrm{ref}}(y^{+}\mid x)}
-
\log\frac{\pi_{\theta}(y^{-}\mid x)}
{\pi_{\mathrm{ref}}(y^{-}\mid x)}
\right]
\right)
\right],
\end{equation}

where $\beta$ penalizes the deviation from the reference policy. The objective therefore encourages a larger margin between the policy's relative likelihood of the chosen and rejected responses.

For an autoregressive language model, the log-likelihood of the rejected response decomposes as a sum over its tokens. Standard DPO implicitly assigns the same contribution to all tokens in this sum, regardless of whether a particular token is actually responsible for the response being rejected. We instead associate each rejected token $y^{-}_t$ with a weight $w_t$ and replace the rejected sequence-level log-ratio with a weighted sum of per-token log-ratios:

\begin{equation}
\log
\frac{\pi_{\theta}(y^{-}\mid x)}
{\pi_{\mathrm{ref}}(y^{-}\mid x)}
\quad\longrightarrow\quad
\sum_{t=1}^{|y^{-}|}
w_t
\left[
\log \pi_{\theta}(y^{-}_t\mid x,y^{-}_{<t})
-
\log \pi_{\mathrm{ref}}(y^{-}_t\mid x,y^{-}_{<t})
\right].
\end{equation}

The resulting weights control how strongly individual rejected tokens contribute to the negative preference signal. Tokens assigned small weights are only weakly penalized, whereas tokens assigned large weights retain a stronger contribution to the DPO objective. We constrain each weight to lie in $[0,1]$, where $w_t=1$ recovers the contribution assigned to that token under standard DPO. Thus, our method does not amplify the penalty of any rejected token beyond the standard DPO objective; instead, it selectively downweights tokens whose penalization is estimated to interfere with preferred behavior. The relation with the standard DPO objective is detailed in Appendix \ref{app:relation_dpo}.

\subsection{Gradient-Based Token Weights}

Our goal is to assign gradient-based token weights according to the effect that penalizing each rejected token would have on learning the chosen response. To characterize this interaction, we consider the parameter-space update induced by each token. For a rejected token $y^{-}_t$, we compute the gradient of its conditional log-likelihood with respect to the model parameters:

\begin{equation}
g^{-}_t
=
\nabla_{\theta}
\log \pi_{\theta}(y^{-}_t\mid x,y^{-}_{<t}).
\end{equation}

This gradient represents the direction in parameter space that would increase the probability of the rejected token. Consequently, penalizing the token induces an update in the opposite direction. We compare this token-level gradient with a gradient representing the behavior encouraged by the chosen response. In the simplest setting, this gradient is computed using the complete chosen response:

\begin{equation}
g^{+}
=
\nabla_{\theta}
\log \pi_{\theta}(y^{+}\mid x).
\end{equation}

For each rejected token, we compute two similarity scores against different reference directions. As the first source, we compare its gradient $g^{-}_t$ with the gradient associated with its corresponding chosen response, denoted by $g^{+}_{\mathrm{pair}}$. This captures whether penalizing the rejected token would interfere with learning the preferred response from the same preference pair. In terms of the second source, we compare $g^{-}_t$ with an aggregate chosen gradient averaged across all $N=|\mathcal{D}|$ preference pairs, $g^{+}_{\mathrm{global}}=\frac{1}{N}\sum_{i=1}^{N} g^{+}_{\mathrm{pair},i}$. This second similarity captures whether penalizing the token would interfere with parameter directions that are consistently supported by preferred responses more broadly. We denote these two similarities as

\begin{equation}
s^{\mathrm{pair}}_t
=
\operatorname{sim}(g^{-}_t,g^{+}_{\mathrm{pair}})
\qquad\text{and}\qquad
s^{\mathrm{global}}_t
=
\operatorname{sim}(g^{-}_t,g^{+}_{\mathrm{global}}).
\end{equation}

Strong alignment with either source is sufficient to indicate that penalizing the token may suppress useful preferred behavior. As such, we take the maximum of the two similarities for each rejected token to ensure a conservative criterion:

\begin{equation}
s_t
=
\max
\left(
s^{\mathrm{pair}}_t,
s^{\mathrm{global}}_t
\right).
\end{equation}

This choice reflects two complementary notions of interference. A rejected token receives a weaker penalty if suppressing it would conflict either with its paired chosen response or with behavior that is consistently supported by chosen responses across the dataset. If the resulting similarity is high, the rejected token and at least one of the chosen signals induce similar parameter updates. Penalizing that token would therefore move the model in the opposite direction and may interfere with preferred behavior, so we assign it a smaller weight. Conversely, when the rejected-token gradient is dissimilar to both its paired chosen gradient and the aggregate chosen gradient, suppressing the token is less likely to interfere with preferred behavior, and it receives a larger loss weight.

Finally, we transform $s_t$ into a token weight $w_t \in [0,1]$ using a monotonically decreasing function. Weights close to $1$ approach the original contribution of the token under standard DPO, while smaller values progressively reduce its negative contribution. In this manner, our weighting mechanism protects both preference-specific information from the paired chosen response and more general preferred behavior across the entire dataset.

\subsection{Experimental and Implementation Decisions}

Computing a separate token gradient over all model parameters is prohibitively expensive for modern LLMs. We therefore measure gradient alignment in the parameter space of a rank-64 PiSSA-initialized adapter~\citep{meng2024pissa} attached to the frozen base model. The adapter is used only as a \emph{gradient probe}; the subsequent DPO stage performs full-model fine-tuning. Because PiSSA initializes the low-rank factors from the principal singular components of each weight matrix, the adapter gradients are projections of the full-parameter gradients onto these principal subspaces (see Appendix~\ref{app:pissa}). A property reported by \citet{meng2024pissa} motivates this choice: the training loss and gradient-norm trajectories of PiSSA follow those of full-parameter fine-tuning closely, and do so consistently across ranks from $1$ to $128$ on several models. This indicates that gradients restricted to the principal subspaces preserve most of the full-parameter gradient signal.

To construct the chosen direction, we aggregate the token-level gradients of the chosen completion while compensating for differences in their magnitude. For cross-entropy, the gradient of $\log p_i^{+}$ with respect to the logits has $L_1$ norm $2(1-p_i^{+})$, where $p_i^{+}$ is the probability assigned to the target token. As such, tokens that the model already predicts confidently contribute close to nothing to an unweighted sum, and the aggregate direction is determined by the few tokens the model finds surprising. We instead weight each chosen-token gradient by the inverse of its logit-level $L_1$ norm,

\begin{equation}
\label{eq:chosen_direction}
g^{+}
=
\sum_{i=1}^{T^{+}}
\alpha_i g_i^{+},
\qquad
\alpha_i
=
\frac{1}{2(1-p_i^{+})},
\end{equation}

with $\alpha_i$ treated as a constant (detached from the computational graph). The result approximates an unweighted average over token gradient directions rather than a magnitude-weighted one, which is the intended reference for the alignment scores: the direction should reflect what the chosen response as a whole supports, not what its most surprising tokens support. The derivation and the numerical stabilization are given in Appendix~\ref{app:chosen_direction}.

We use the \textit{dot product} to measure gradient alignment. 
The dot product is particularly convenient computationally: once $g_t^{-}$ is obtained, its projections onto multiple reference directions can be computed directly, and only the resulting scalar products need to be stored. We therefore never materialize the full token-by-parameter gradient matrix. Rejected tokens are processed in chunks to further bound memory usage. More details can be found in Appendix \ref{app:efficient_alignment_computation}.


Finally, we convert the alignment scores into rejected-token weights. Since the raw dot products scale with the gradient norms and these vary considerably across examples, we normalize the scores of each rejected response by their largest absolute value and apply a sigmoid to the negated result:
\begin{equation}
\label{eq:weight}
s_t =
\max\left(
\left\langle g_t^{-}, g_{\mathrm{pair}}^{+}\right\rangle,
\left\langle g_t^{-}, g_{\mathrm{global}}^{+}\right\rangle
\right),
\qquad
w_t
=
\sigma\!\left(-\frac{s_t}{\max_{t'} |s_{t'}|}\right),
\end{equation}
where the maximum runs over all tokens of the same rejected response.

The $\frac{s_t}{\max_{t'} |s_{t'}|}$ mapping is symmetric around $0$ because the score is signed and both signs are informative: \textit{positive} alignment marks a token whose suppression would interfere with preferred behavior, \textit{negative} alignment marks a conflicting token whose suppression moves the model toward the preferred update, while score close to zero denote \textit{orthogonal} tokens who are separated from conflicting ones rather than collapsed onto the same weight. The weights attenuate the rejected penalty at individual positions, and since no weight exceeds $1$, no rejected token is penalized more strongly than under standard DPO. Highly aligned tokens receive smaller weights, while conflicting tokens receive weights closer to $1$ and therefore retain more of the standard DPO penalty. Only rejected tokens are reweighted; chosen-token weights remain fixed to $1$. Appendix~\ref{app:qualitative} shows the resulting weights on concrete examples.

\subsection{Algorithm and Training Procedure}

Algorithm~\ref{alg:gradient_weighted_dpo} summarizes the complete procedure. We first precompute rejected-token weights using gradient alignment in the frozen PiSSA probe space, then keep these weights fixed during full-model DPO training. We use $G^{+}(x,y^{+})$ to denote the chosen-response gradient direction defined in Equation~\ref{eq:chosen_direction}; each rejected token is compared against the pair-specific and dataset-level chosen directions, with stronger alignment yielding a smaller DPO penalty.

\begin{algorithm}[ht]
\small
\caption{GAW-PO: Gradient-Aligned Token Reweighting for DPO}
\label{alg:gradient_weighted_dpo}
\begin{algorithmic}[1]
\Require Preference dataset $\mathcal{D}=\{(x_i,y_i^{+},y_i^{-})\}_{i=1}^{N}$, SFT policy $\pi_0$, reference policy $\pi_{\mathrm{ref}}$
\Ensure Trained policy $\pi_{\theta}$
\State Attach a frozen-base PiSSA probe with parameters $\phi$ to $\pi_0$.
\Comment{Defines the gradient space}
\Statex \textbf{Gradient-based weight precomputation}
\State $g_{\mathrm{pair},i}^{+} \gets G^{+}(x_i,y_i^{+})$ for $i=1,\ldots,N$
\Comment{Preferred direction of each pair}
\State $g_{\mathrm{global}}^{+} \gets \frac{1}{N}\sum_{i=1}^{N} g_{\mathrm{pair},i}^{+}$
\Comment{Dataset-level preferred direction}
\For{$i=1,\ldots,N$}
\For{$t=1,\ldots,|y_i^{-}|$}
\Comment{Alignment scores for each token $t$}
\State $g_t^{-}\gets\nabla_{\phi}\log\pi_{\phi}(y_{i,t}^{-}\mid x_i,y_{i,<t}^{-})$
\State $s_t\gets\max\left(\langle g_t^{-},g_{\mathrm{pair},i}^{+}\rangle,\langle g_t^{-},g_{\mathrm{global}}^{+}\rangle\right)$
\Comment{Protect local or global preferred behavior}
\EndFor
\State $m \gets \max_{t}\, |s_t|$
\Comment{Largest absolute score of this response}
\For{$t=1,\ldots,|y_i^{-}|$}
\Comment{Token weights}
\State $w_{i,t} \gets \sigma(-s_t/m)$
\Comment{Higher alignment $\Rightarrow$ weaker penalty}
\EndFor
\EndFor
\Statex \textbf{Weighted DPO training}
\State Initialize $\pi_{\theta}$.
\While{training}
\State Sample $i$; compute the chosen log-ratio $r^{+}$ with uniform token weights.
\State $r^{-} \gets \sum_t w_{i,t}\left[\log\pi_{\theta}(y_{i,t}^{-}\mid x_i,y_{i,<t}^{-})-\log\pi_{\mathrm{ref}}(y_{i,t}^{-}\mid x_i,y_{i,<t}^{-})\right]$
\State Update $\theta$ using $\mathcal{L}=-\log\sigma\left(\beta(r^{+}-r^{-})\right)$.
\EndWhile
\State \Return $\pi_{\theta}$
\end{algorithmic}
\end{algorithm}

\section{Experimental Setup}
\label{sec:experimental-setup}

\subsection{Experimental Details}
\label{sec:hyperparameters}

We use \textit{Olmo-3-7B-Instruct-SFT}\footnote{\url{https://huggingface.co/allenai/Olmo-3-7B-Instruct-SFT}}, the supervised fine-tuning (SFT) checkpoint of the 7B OLMo 3 Instruct model~\citep{olmo2025olmo3}, as the initialization for all experiments. OLMo 3 is a fully open model family, with publicly available model weights, intermediate checkpoints, and training data, making it suitable for controlled studies of optimization dynamics.

For preference training, we use a filtered subset of \textit{Dolci-Instruct-DPO}\footnote{\url{https://huggingface.co/datasets/allenai/Dolci-Instruct-DPO}}, the preference-training dataset used for the OLMo 3 Instruct DPO stage~\citep{olmo2025olmo3}. For a controlled setting, we retain only English, single-turn user-assistant preference pairs, and restrict the total textual content to at most 2048 tokens. The resulting dataset contains approximately 209K preference pairs (80\% of the original dataset). We use the standard OLMo 3 chat template and mask prompt tokens when computing completion losses.

For our method, we perform full fine-tuning for one epoch with a maximum sequence length of 2048, learning rate $1\times10^{-5}$, effective batch size 128, $\beta=0.1$, and warmup ratio $0.1$. We use a linear learning-rate schedule, 8-bit AdamW, \texttt{bfloat16}. Unless stated otherwise for a specific ablation or experiment, this constitutes our default experimental setup. For gradient-weight computation, we attach a rank-64 PiSSA adapter with $\alpha=128$, zero dropout, and all linear layers as target modules. The adapter is used exclusively as a gradient probe; preference optimization itself remains a full-model fine-tune.

Precomputing the alignment scores over all the preference pairs costs approximately 63 GPU-hours on H200 GPUs: one reverse-mode pass per example to obtain the chosen direction, followed by two forward-mode (JVP) passes for the paired and dataset-level projections (see Appendix~\ref{app:efficient_alignment_computation}). The precomputation is performed once per dataset and is reused across every training run in our experiments. For comparison, a single full-model training run takes approximately 40 GPU-hours. 

\subsection{Baselines}
\label{sec:baselines}

We compare against the starting SFT checkpoint, standard DPO~\citep{rafailov2023dpo}, 3 sequence-level variants (i.e., IPO~\citep{azar2024ipo}, KTO~\citep{ethayarajh2024kto} and SimPO \citep{meng2024simpo}), and 3 token-weighting methods (i.e., TIS-DPO~\citep{liu2025tis}, TI-DPO \citep{yang2026tidpo} and AAO~\citep{li2025aao}). All methods start from the same SFT checkpoint and are trained for one epoch on the same preference data. Method-specific hyperparameters follow the original papers or their official implementations. Appendix~\ref{app:hparams} lists all values and the few deviations.

\subsection{Evaluation Benchmarks}
\label{sec:benchmarks}

We evaluate all models using a subset of the OLMES benchmark suite~\citep{gu2025olmes}, following the same evaluation framework used in the OLMo 3 paper~\citep{olmo2025olmo3}. All results are reported across 3 runs with different seeds. We use the official OLMES evaluation recipes for OLMo 3 and select benchmarks spanning four broad capability groups: \textbf{Mathematics:} AIME 2024 \citep{aime24} and 2025 \citep{aime25}, MATH \citep{hendrycks2021math}, and Omega 500 \citep{sun2024omega}; \textbf{Reasoning:} AGI Eval English \citep{zhong2024agieval} and BigBenchHard-BBH \citep{suzgun2023bbh} with chain-of-thought prompting; \textbf{Code:} HumanEval+ \citep{chen2021codex, liu2023evalplus}, LiveCodeBench \citep{jain2025livecodebench}, and MBPP+ \citep{austin2021mbpp, liu2023evalplus}; \textbf{Question Answering:} GPQA \citep{rein2024gpqa} and MMLU \citep{hendrycks2021mmlu} with chain-of-thought prompting.

\section{Results}

\subsection{Comparison with Existing Methods}

Table~\ref{tab:benchmark_results} compares GAW-PO against standard DPO and existing preference-optimization and token-weighting approaches using the experimental setup from Section~\ref{sec:experimental-setup}. Our method achieves the strongest overall performance, with an average score of \textbf{41.99}. This corresponds to an absolute improvement of \textbf{0.97 points} over standard DPO (41.02) and \textbf{0.65 points} over TIS-DPO (41.34), the strongest competing method on average. Moreover, the gains are distributed across several capability groups; GAW-PO obtains the best result on 8 of the 11 evaluated benchmarks.

The largest improvements are observed on reasoning and coding tasks. On BBH, our method reaches \textbf{49.81}, improving over standard DPO by 4.36 points and over the strongest competing baseline by 4.31 points. Similarly, on LiveCodeBench, we obtain \textbf{24.78}, compared with 21.36 for DPO and 21.88 for TIS-DPO. We also observe consistent gains on HumanEval+ (73.52), MATH (70.35), AGI Eval (61.15), MBPP+ (57.73), and MMLU (67.92). These results indicate that reducing harmful rejected-token updates can improve performance across different downstream tasks. Appendix~\ref{app:qualitative} shows that the weights fall on the tokens where a rejected response diverges from the chosen one, sparing correct content.

The improvement is not universal across all benchmarks. TIS-DPO performs best on AIME 2025 and GPQA, while IPO obtains the highest score on Omega 500. In particular, our method underperforms on GPQA and Omega 500. Nevertheless, the stronger performance across the majority of benchmarks results in the highest aggregate score among all evaluated methods.

Overall, these results support our hypothesis that token weights should depend on their \emph{interaction with preferred update directions}, rather than only on token-level importance in isolation. By downweighting rejected tokens whose gradients are compatible with preferred behavior and concentrating the negative signal on conflicting updates, our method produces a more targeted preference-optimization signal and improves overall downstream performance.

\begin{table*}[ht]
    \centering
    \small
    \setlength{\tabcolsep}{4.2pt}
    \caption{
        Performance comparison across evaluation benchmarks: mean performance (\%) across 3 runs, standard deviation shown in gray.
        Best results per benchmark are \textbf{bolded}.
    }
    \label{tab:benchmark_results}
    \resizebox{\textwidth}{!}{
    \begin{tabular}{lccccccccc}
        \toprule
        \textbf{Benchmark}
        & \textbf{SFT}
        & \textbf{DPO}
        & \textbf{IPO}
        & \textbf{KTO}
        & \textbf{SimPO}
        & \textbf{TIS-DPO}
        & \textbf{TI-DPO}
        & \textbf{AAO}
        & \textbf{GAW-PO (Ours)} \\
        \midrule

        AIME 2024
        & 5.73 \std{0.55}
        & 6.28 \std{0.80}
        & 6.25 \std{0.21}
        & 5.66 \std{0.97}
        & 6.63 \std{0.33}
        & 6.46 \std{0.81}
        & 6.53 \std{0.51}
        & 6.67 \std{0.38}
        & \textbf{7.36} \std{0.52} \\

        AIME 2025
        & 6.88 \std{1.04}
        & 8.33 \std{1.00}
        & 7.57 \std{0.66}
        & 7.81 \std{0.65}
        & 7.95 \std{0.69}
        & \textbf{8.47} \std{1.16}
        & 8.44 \std{0.99}
        & 7.88 \std{0.99}
        & 7.95 \std{0.42} \\

        MATH
        & 67.70 \std{0.27}
        & 68.69 \std{0.38}
        & 68.07 \std{0.73}
        & 67.77 \std{0.47}
        & 68.61 \std{0.81}
        & 69.41 \std{0.36}
        & 69.30 \std{0.47}
        & 68.59 \std{0.37}
        & \textbf{70.35} \std{0.24} \\

        Omega 500
        & 11.53 \std{0.76}
        & 12.27 \std{1.03}
        & \textbf{12.47} \std{0.95}
        & 11.87 \std{1.85}
        & 11.93 \std{0.95}
        & 12.33 \std{0.92}
        & 12.13 \std{0.76}
        & 11.93 \std{0.99}
        & 11.60 \std{1.31} \\

        \midrule

        AGI Eval
        & 58.51 \std{0.58}
        & 60.31 \std{1.08}
        & 59.45 \std{0.62}
        & 59.46 \std{0.59}
        & 58.47 \std{1.40}
        & 60.12 \std{0.62}
        & 59.75 \std{0.34}
        & 59.14 \std{0.29}
        & \textbf{61.15} \std{0.74} \\

        BBH (CoT)
        & 44.40 \std{0.77}
        & 45.45 \std{0.55}
        & 45.37 \std{0.85}
        & 45.12 \std{0.09}
        & 45.50 \std{0.64}
        & 45.33 \std{0.11}
        & 45.13 \std{0.82}
        & 44.87 \std{0.69}
        & \textbf{49.81} \std{0.21} \\

        \midrule

        HumanEval+
        & 68.84 \std{0.48}
        & 71.63 \std{0.94}
        & 71.12 \std{0.58}
        & 69.51 \std{0.44}
        & 70.98 \std{0.79}
        & 71.87 \std{0.27}
        & 71.59 \std{1.06}
        & 71.32 \std{0.90}
        & \textbf{73.52} \std{0.41} \\

        LiveCodeBench
        & 20.30 \std{0.34}
        & 21.36 \std{0.26}
        & 20.72 \std{0.10}
        & 20.72 \std{0.14}
        & 21.13 \std{0.13}
        & 21.88 \std{0.25}
        & 21.24 \std{0.21}
        & 20.92 \std{0.35}
        & \textbf{24.78} \std{0.37} \\

        MBPP+
        & 56.38 \std{0.37}
        & 56.71 \std{0.31}
        & 56.78 \std{0.33}
        & 56.49 \std{0.12}
        & 56.67 \std{0.27}
        & 57.13 \std{0.68}
        & 56.71 \std{0.19}
        & 56.45 \std{0.31}
        & \textbf{57.73} \std{1.17} \\

        \midrule

        GPQA
        & 31.47 \std{1.74}
        & 33.26 \std{1.36}
        & 31.99 \std{2.68}
        & 31.92 \std{1.61}
        & 32.14 \std{0.67}
        & \textbf{34.08} \std{2.41}
        & 31.55 \std{2.41}
        & 30.88 \std{0.52}
        & 29.76 \std{0.26} \\

        MMLU (CoT)
        & 66.75 \std{0.32}
        & 66.91 \std{0.49}
        & 67.48 \std{0.17}
        & 66.86 \std{0.11}
        & 66.92 \std{0.59}
        & 67.70 \std{0.19}
        & 67.63 \std{0.28}
        & 67.09 \std{0.19}
        & \textbf{67.92} \std{0.25} \\

        \midrule
        \textbf{Average}
        & 39.86
        & 41.02
        & 40.66
        & 40.29
        & 40.63
        & 41.34
        & 40.91
        & 40.52
        & \textbf{41.99} \\

        \bottomrule
    \end{tabular}
    }
\end{table*}

\subsection{Robustness to the DPO Regularization Strength}
\label{sec:beta_robustness}
    
The DPO parameter $\beta$ (see Equation~\ref{eq:dpo_formula}) controls the strength of the implicit regularization toward the reference policy. Larger values constrain the optimized policy more strongly, whereas smaller values permit larger deviations from the reference model. In practice, DPO is commonly trained with moderate values of $\beta\approx 0.1$, since aggressive preference optimization can lead to substantial degradation due to policy collapse. Under our motivating hypothesis, part of this degradation may arise because stronger optimization also amplifies the penalty applied to locally correct tokens in rejected responses, suppressing both useful behavior and genuinely undesirable parts of the response.

\begin{figure*}[ht]
    \centering
    \includegraphics[width=0.55\textwidth]{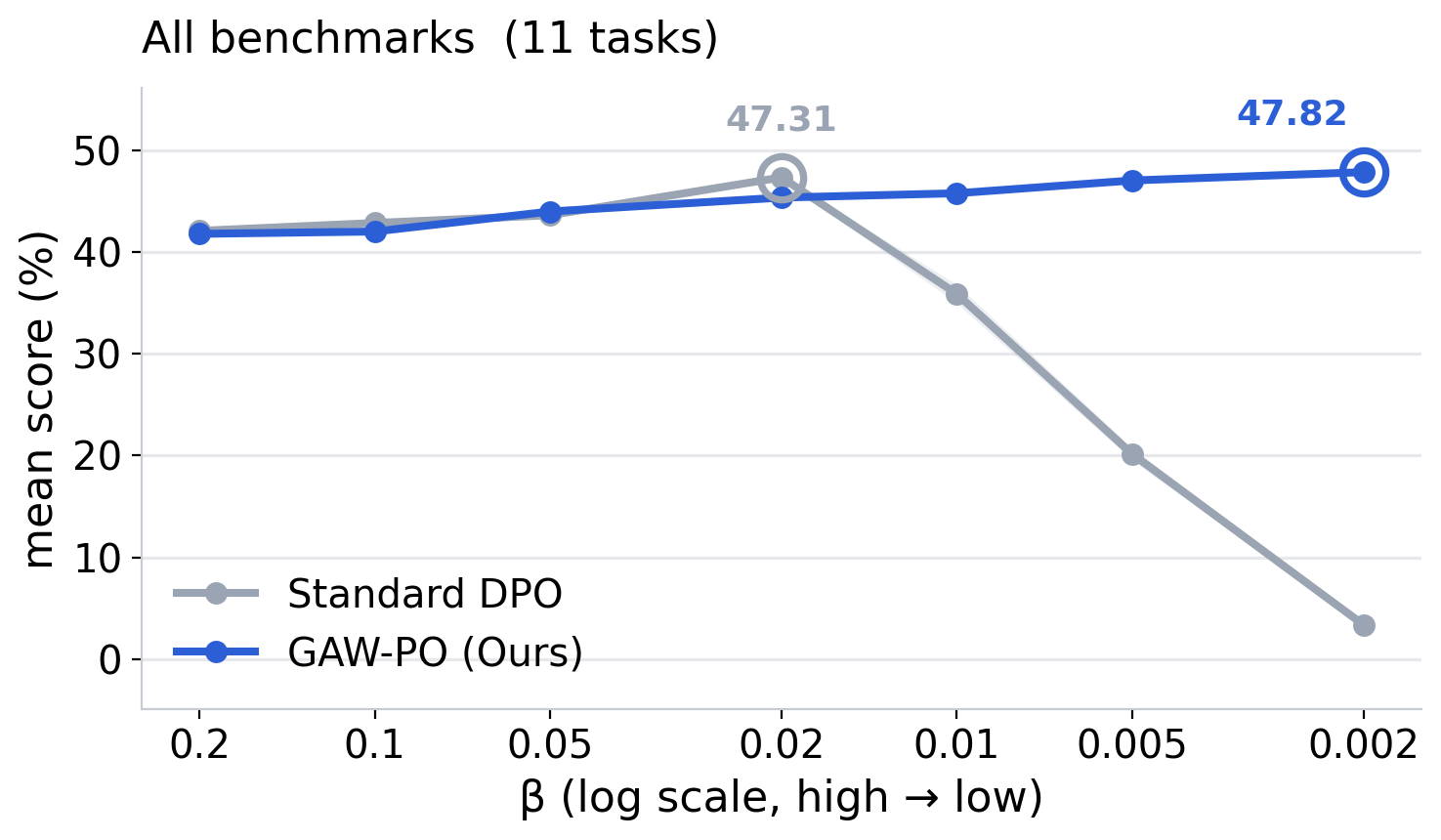}
    \caption{
    Average benchmark performance for different values of the DPO regularization parameter $\beta$.
    Standard DPO exhibits a narrow optimum around $\beta=0.02$ and rapidly degrades under more aggressive optimization, whereas our method remains stable and continues to improve as $\beta$ decreases.
    }
    \label{fig:all_results}
\end{figure*}

We investigate whether our token-level weighting improves robustness in this regime by sweeping
$\beta \in \{0.2, 0.1, 0.05, 0.02, 0.01, 0.005, 0.002\}$,
using the same learning rate of $1\times10^{-5}$ for both standard DPO and GAW-PO. Figure~\ref{fig:all_results} reports the average score across all 11 evaluation benchmarks.

The two methods behave similarly in the strongly regularized regime. Standard DPO reaches its best performance around $\beta=0.02$, but deteriorates sharply once $\beta$ is decreased further. In contrast, our method remains stable over the entire sweep and improves as the regularization is relaxed. The difference becomes pronounced at $\beta=0.01$ and grows rapidly for $\beta=0.005$ and $\beta=0.002$. At the most aggressive setting, standard DPO loses almost all of its performance, whereas GAW-PO achieves its best aggregate result. Notably, GAW-PO at $\beta=0.002$ also surpasses the best standard-DPO configuration observed anywhere in the sweep. The category-level results detailed in Appendix \ref{app:beta_sweep_results} reveal the same pattern across mathematics, reasoning, coding, and QA. 

These results indicate that our weighting scheme substantially enlarges the range of $\beta$ values over which preference optimization remains effective. Standard DPO requires relatively strong regularization to prevent destructive updates, resulting in a narrow range of effective optimization strengths. By reducing the contribution of rejected tokens whose gradients are aligned with preferred behavior, GAW-PO permits substantially larger deviations from the reference policy without loss in downstream capability. This is consistent with our hypothesis that part of the instability of aggressive DPO originates from negative updates applied to behavior that should be preserved. Random or uniform weights with the same mean lose relative to gradient-aligned weights, so the gain is not due to reduced penalty magnitude alone (see Appendix~\ref{app:controls}).

\subsection{Multiple Preferred Update Directions}

Our main method compares each rejected-token gradient against both its paired chosen gradient and a single aggregate gradient over all chosen responses. However, a single dataset-level mean may be too coarse for a heterogeneous preference dataset, where different types of preferred behavior can induce distinct update directions. We therefore investigate a clustered variant in which chosen gradients are partitioned into $K$ groups in the parameter space. For each cluster, we compute a centroid representing its average preferred update. The score of a rejected token becomes
$
s_t^{\mathrm{cluster}}
=
\max\left(
    \left\langle g_t^{-}, g_{\mathrm{pair}}^{+}\right\rangle,
    \max_{k=1,\ldots,K}
    \left\langle g_t^{-}, c_k^{+}\right\rangle
\right),
$
where $c_k^{+}$ denotes the chosen-gradient centroid of cluster $k$. Thus, a rejected token is protected if it aligns either with its paired chosen response or with any recurring preferred direction in the dataset.

\begin{table}[t]
    \centering
    \footnotesize
    \setlength{\tabcolsep}{4pt}
    \caption{
    Average performance for paired+global and clustered preferred directions at $\beta=0.002$.
    We report mean performance across 3 runs.
    Full per-benchmark results are in Appendix~\ref{app:cluster_results}.
    }
    \label{tab:cluster_results}
    \begin{tabular}{@{}lcccc@{}}
        \toprule
        \textbf{Metric}
        & \textbf{Mean}
        & \textbf{$K$=2}
        & \textbf{$K$=4}
        & \textbf{$K$=8} \\
        \midrule
        \textbf{Average}
        & 47.82 
        & 46.60 
        & \textbf{47.94} 
        & 47.89  \\
        \bottomrule
    \end{tabular}
    \vspace*{-0.5em}
\end{table}

Among the clustered variants, $K=4$ performs best, reaching \textbf{47.94} compared with 47.82 for the original paired+global formulation. $K=8$ also slightly improves over the global mean, while $K=2$ performs worse. The gains for $K=4$ are most visible in question answering and mathematics, suggesting that multiple preferred directions can be beneficial, although performance does not improve monotonically with $K$. Additional clustering details and full per-benchmark results are provided in Appendix~\ref{app:clustered_directions} and \ref{app:cluster_results}.

\section{Conclusion and Future Work}
\label{app:conclusion}

We introduced a gradient-aligned token reweighting approach for DPO that accounts for the effect of penalizing individual rejected tokens on preferred behavior. Rather than estimating token importance in isolation, GAW-PO compares rejected-token gradients with preferred update directions and reduces the penalty on tokens whose suppression would interfere with learning the chosen response. We make this computation practical through a low-rank PiSSA gradient probe, while retaining full-model preference optimization during training.

Across our evaluation suite, the proposed method achieves the highest average performance among the considered preference-optimization methods, outperforming standard DPO and the strongest competing baseline. More importantly, the method remains effective under substantially more aggressive preference optimization: as $\beta$ decreases, standard DPO degrades sharply, whereas GAW-PO remained stable and often improved. We additionally explored representing preferred behavior through multiple clustered update directions. While this provides modest gains for some values of $K$, the results also showed that increasing the number of preferred directions is not sufficient by itself; the structure and quality of these directions are also important.

In terms of future work, we plan to extend our method beyond pairwise objectives such as DPO to online reinforcement-learning methods, particularly Group Relative Policy Optimization (GRPO) \citep{shao2024deepseekmath}. Although GRPO assigns updates based on relative rewards across multiple sampled responses, low-reward trajectories may still contain useful tokens or reasoning steps that are also supported by higher-reward responses. Our gradient-alignment framework could provide token-level credit assignment by downweighting negative token updates whose gradients align with high-advantage responses, while preserving stronger penalties for conflicting tokens.

Finally, Appendix~\ref{app:limitations} discusses the study's limitations, from the cost of the offline weight computation to the scope of our evaluation, each pointing to a direction for future work.

\section*{AI Use Disclosure}
LLMs were used as assistive tools in the preparation of this manuscript and the accompanying code. Specifically, we used ChatGPT 5.0+ (OpenAI, \footnote{\url{https://chatgpt.com/}}) and Claude Opus 5.0+ (Anthropic, \footnote{\url{https://claude.ai/}}) to assist with drafting and revising portions of the manuscript. Their use included grammar and spell checking, improving clarity and phrasing, reducing repetition, and general language polishing. Claude was additionally used to assist with suggesting and debugging portions of the experimental code, as well as with the generation and refinement of the teaser figure used in the manuscript. LLMs were used only as assistive tools; all generated text and code suggestions were reviewed, verified, and, where necessary, modified by the authors. The authors remain fully responsible for the content of the manuscript, the implementation, the experimental methodology, and the reported results.

\bibliography{iclr2027_conference}
\bibliographystyle{iclr2027_conference}

\appendix

\section{Additional Gradient-Weighting Details}
\label{app:gradient_details}

\subsection{Relation to the DPO derivation.}
\label{app:relation_dpo}

Standard DPO is derived as an exact reparameterization of the KL-constrained sequence-level bandit objective, where the implicit reward $r_\theta(x,y)=\beta\log\frac{\pi_\theta(y\mid x)}{\pi_{\mathrm{ref}}(y\mid x)}$ is given by the log-ratio of two normalized distributions~\citep{rafailov2023dpo}. The weighted rejected term $r_w(x,y^{-})=\beta\sum_t w_t\big[\log\pi_\theta(y^-_t\mid x,y^-_{<t})-\log\pi_{\mathrm{ref}}(y^-_t\mid x,y^-_{<t})\big]$ does not have this form. Therefore, the weighted learning objective should be viewed as a contrastive surrogate rather than as the exact dual of a KL-constrained sequence-level objective.

However, its impact on optimization is straightforward since the weights are detached constants,
\begin{equation}
\nabla_\theta\mathcal{L}
=
-\beta\,\mathbb{E}\Big[
\sigma\big(-\beta(r^{+}-r_w^{-})\big)
\Big(
\nabla_\theta\log\pi_\theta(y^{+}\mid x)
-\sum_{t}w_t\,\nabla_\theta\log\pi_\theta(y^{-}_t\mid x,y^{-}_{<t})
\Big)\Big],
\end{equation}

is simply the standard DPO gradient with the gradient contribution of each rejected token scaled by $w_t$. The method can therefore be understood as applying token-level credit assignment to the DPO update.

Consequently, $w_t$ can be viewed as inducing a token-dependent regularization coefficient $\beta w_t$ for the rejected response. This interpretation is not an exact derivation, because the coefficient is not defined consistently as a function of state across both chosen and rejected trajectories. Nevertheless, it places our objective in the same general family as token-level variants such as TDPO~\citep{zeng2024tdpo} and TIS-DPO~\citep{liu2025tis}, which likewise depart from the sequence-level dual in order to obtain finer-grained credit assignment.

\subsection{PiSSA Gradient Probe}
\label{app:pissa}

Computing a separate gradient for every rejected token over all parameters of a 7B model is prohibitively expensive. We therefore measure gradient alignment in a low-rank parameter space defined by a PiSSA-initialized adapter~\citep{meng2024pissa} attached to the frozen SFT model. Importantly, this adapter is used only as a \emph{gradient probe}: it is never optimized, and the subsequent preference-optimization stage remains a full-model fine-tune.

PiSSA initializes the two low-rank factors from a truncated singular-value decomposition of each target weight matrix. For
\begin{equation}
W = U\Sigma V^\top,
\end{equation}
the rank-$r$ component is represented using
\begin{equation}
A = \sqrt{\Sigma_r}V_r^\top,
\qquad
B = U_r\sqrt{\Sigma_r},
\end{equation}
while the corresponding contribution is subtracted from the frozen base weight. Consequently, the function represented by the model is unchanged at initialization, while both low-rank factors contain non-zero values and admit informative gradients.

In all experiments, we use a rank-$64$ PiSSA adapter with scaling parameter $\alpha=128$, zero dropout, and all linear layers as target modules. The same frozen probe is used to compute the chosen-response directions and rejected-token gradients.

\subsection{Construction of the Chosen Direction}
\label{app:chosen_direction}

For a chosen token $y_i^{+}$ with logits $z_i$ and target probability $p_i^{+}=\operatorname{softmax}(z_i)_{y_i^{+}}$, the gradient of the token log-likelihood with respect to the logits is
\begin{equation}
\frac{\partial \log p_i^{+}}{\partial z_i}
=
\mathrm{onehot}(y_i^{+})-\operatorname{softmax}(z_i),
\end{equation}
whose $L_1$ norm is $(1-p_i^{+})+\sum_{j\neq y_i^{+}} p_{i,j} = 2(1-p_i^{+})$. The parameter gradient is $g_i^{+} = (\partial z_i/\partial\phi)^{\top}\,\partial\log p_i^{+}/\partial z_i$. The weight $\alpha_i = 1/(2(1-p_i^{+}))$ in Equation~\ref{eq:chosen_direction} therefore rescales each token so that its logit-level error vector has unit $L_1$ norm. It does not normalize the parameter gradient itself.

Without the rescaling, $\|\partial\log p_i^{+}/\partial z_i\|_1 = 2(1-p_i^{+})$ vanishes for tokens the model already predicts with high confidence, and the sum $\sum_i g_i^{+}$ is dominated by the small number of tokens with low $p_i^{+}$. Since $g^{+}$ serves as the reference against which every rejected token is scored, this would make the alignment scores depend on the most surprising chosen tokens rather than on the chosen response as a whole. The rescaled sum instead approximates an unweighted average over token gradient directions.

\subsection{Efficient Alignment Projection Computation}
\label{app:efficient_alignment_computation}

Explicitly computing and storing a gradient for every rejected token would require a matrix with dimensions
$T^{-}\times D$, where $D\approx1.6\times10^8$ in our PiSSA probe. However, our scoring function never requires the full rejected-token gradients themselves; it only requires their inner products with a small number of preferred directions.

We therefore compute these quantities directly using forward-mode automatic differentiation. Let

\begin{equation}
f(\phi)
=
\left[
\log \pi_\phi(y_1^{-}\mid x),
\ldots,
\log \pi_\phi(y_{T^-}^{-}\mid x,y_{<T^-}^{-})
\right]
\end{equation}

denote the vector of rejected-token log-likelihoods. For any fixed preferred direction $v$, a Jacobian vector product gives

\begin{equation}
J_f(\phi)v
=
\left[
\left\langle g_1^{-},v\right\rangle,
\ldots,
\left\langle g_{T^-}^{-},v\right\rangle
\right],
\end{equation}

where $g_t^{-}=\nabla_\phi \log\pi_\phi(y_t^{-}\mid x,y_{<t}^{-})$. Thus, a single directional pass computes the alignment scores for all rejected tokens simultaneously.

For the main method, we evaluate this projection for the pair-specific chosen direction $g_{\mathrm{pair}}^{+}$ and the dataset-level direction $g_{\mathrm{global}}^{+}$, and then take their element-wise maximum. Consequently, we never materialize the full $T^{-}\times D$ rejected-gradient matrix; only the resulting scalar alignment scores are stored. The number of differentiation passes therefore depends on the number of preferred reference directions rather than on the number of rejected tokens.

\section{Multiple Preferred Update Directions}
\label{app:clustered_directions}

\subsection{Clustering Procedure}

The main method represents dataset-level preferred behavior using a single mean chosen gradient. To study whether the chosen-gradient space contains additional structure, we construct multiple preferred reference directions through clustering.

Computing and storing the full $D$-dimensional chosen gradient for every preference example would be impractical. We therefore first project each chosen-response gradient into a $1000$-dimensional CountSketch representation \citep{charikar2002finding}. These sketches are used only for clustering.

We apply spherical $k$-means to the normalized sketch vectors. Spherical clustering groups examples according to gradient direction rather than Euclidean distance or raw gradient magnitude. Given assignments
$\mathcal{C}_1,\ldots,\mathcal{C}_K$, we subsequently recompute each centroid in the original PiSSA parameter space,

\begin{equation}
c_k^{+}
=
\frac{1}{|\mathcal{C}_k|}
\sum_{e\in\mathcal{C}_k}
g_e^{+}.
\end{equation}

The full-space recomputation is necessary because sketch-space centroids cannot be directly compared with token gradients in the PiSSA parameter space.

For a rejected token, the clustered score is

\begin{equation}
s_t^{\mathrm{cluster}}
=
\max
\left(
    \left\langle
    g_t^{-},g_{\mathrm{pair}}^{+}
    \right\rangle,
    \max_{k=1,\ldots,K}
    \left\langle
    g_t^{-},c_k^{+}
    \right\rangle
\right).
\end{equation}

Thus, each rejected token independently selects the preferred direction with which it has the largest alignment. We do not first assign the rejected example to a particular chosen-gradient cluster.

We evaluate $K \in \{2,4,8\}$. All clustered variants are trained at $\beta=0.002$, the most aggressive setting of the sweep in Section~\ref{sec:beta_robustness} and the one at which our method attains its best aggregate result; all other hyperparameters follow the default setup of Section~\ref{sec:hyperparameters}.

\subsection{Efficient Cluster Projections}

The clustered variant uses the same projection mechanism as the main method. The only difference is that the rejected-token gradients are compared against multiple dataset-level preferred directions rather than a single global mean.

For each cluster centroid $c_k^{+}$, we compute

\begin{equation}
J_f(\phi)c_k^{+}
=
\left[
\left\langle g_1^{-},c_k^{+}\right\rangle,
\ldots,
\left\langle g_{T^-}^{-},c_k^{+}\right\rangle
\right],
\end{equation}

using one Jacobian vector product. Repeating this operation for the $K$ cluster centroids yields all token-centroid alignment scores without explicitly constructing any rejected-token gradient vectors. The pair-specific projection is computed in the same way, after which the final score is obtained by taking the element-wise maximum over the paired direction and all cluster directions,

\begin{equation}
s_t^{\mathrm{cluster}}
=
\max\left(
    \left\langle g_t^{-},g_{\mathrm{pair}}^{+}\right\rangle,
    \max_{k=1,\ldots,K}
    \left\langle g_t^{-},c_k^{+}\right\rangle
\right).
\end{equation}

Thus, the clustered extension increases the number of directional passes from two in the main method to $K+1$, while retaining the same memory-efficient computation and avoiding per-token reverse-mode gradient materialization.

\section{Baseline Hyperparameters}
\label{app:hparams}

Following the OLMo~3 7B Instruct DPO recipe \citep{olmo2025olmo3}, all baselines are trained for one epoch with an effective batch size of 128 and a linear learning-rate schedule with 10\% warmup. All baselines use 8-bit AdamW and, unless stated otherwise below, a learning rate of $2\times10^{-6}$.

\paragraph{DPO and IPO.} Both use $\beta = 0.1$. The IPO paper does not give a recipe for LLM training, so we use $\tau = 0.1$, the same value as $\beta$ for DPO.

\paragraph{KTO.} Each preference pair is split into one desirable and one undesirable example, which gives a balanced dataset, so we set $\lambda_D = \lambda_U = 1$ and $\beta = 0.1$. We use a learning rate of $1\times10^{-6}$, the maximum recommended for $\beta = 0.1$ in the TRL implementation of KTO, since higher values degrade training.

\paragraph{SimPO.} We use $\beta = 10$ and $\gamma/\beta = 0.3$ from the official implementation. With its learning rate of $1\times10^{-6}$ the model did not learn in our setup, so we use the shared $2\times10^{-6}$.

\paragraph{TIS-DPO.} We use the variant in which token weights are estimated from two contrastive models obtained with DPO and reverse DPO, with $\mu = \pm 1$ and the log-ratio clamped to $[-0.5, 1.5]$ as in the original paper, and the TDPO$_2$ objective with $\alpha = 0.5$. Weights are normalized per sequence.

\paragraph{TI-DPO.} Token weights mix gradient attribution with a Gaussian position prior, with $\lambda = 0.7$ and $\sigma = T/4$, normalized to mean one per sequence. We use $\beta = 0.1$ and the TDPO$_2$ objective with
$\alpha = 0.5$. Our runs match the \emph{no triplet loss} framework of the original paper. We use the learning rate of $5\times10^{-6}$ from the authors' configuration.

\paragraph{AAO.} We use the smoothing parameter $\alpha = 200$ from the original paper. The paper learns the ambiguity thresholds with a small network; we fix them to $a = 0.6$ and $b = 0.3$.

\section{Qualitative and Control Analyses of Token Weights}
\label{app:weights}

\subsection{Qualitative Examples}
\label{app:qualitative}

Table~\ref{tab:qualitative} shows per-token weights for three rejected completions covering instruction violation, fabricated content, and factual error. Several patterns emerge.

\begin{table}[t]
\centering
\caption{Per-token weights assigned by our method on three rejected completions. Shading is monotone in the weight a token receives (pale $=$ low, deep $=$ high). The weight concentrates on the tokens listed under \textbf{Errors}, while correctly transcribed content receives near-uniform low weight. \newline\scriptsize*The prompts are summarized and example (c) is truncated for readability.}
\label{tab:qualitative}
\small
\setlength{\tabcolsep}{4pt}
\renewcommand{\arraystretch}{1.25}
\begin{tabularx}{\textwidth}{@{}l>{\raggedright\arraybackslash}X@{}}
\toprule
\multicolumn{2}{@{}l@{}}{\textbf{(a) Biomedical relation extraction}} \\[2pt]
\textbf{Prompt} & Extract every drug combination in the sentence \emph{``Reduction of \textbf{ASS1} expression by \textbf{siRNA} significantly sensitized mesothelioma spheroids to the pro-apoptotic effects of \textbf{bortezomib} and of \textbf{cisplatin} plus \textbf{pemetrexed}''}, copying drug names verbatim and labelling each \textsc{pos}, \textsc{neg} or \textsc{comb}. \emph{``In your output, return only the json array and no other text.''} \\[2pt]
\textbf{Chosen} & \texttt{[[\,"bortezomib", "cisplatin", "pemetrexed", "COMB"\,], [\,"cisplatin", "pemetrexed", "COMB"\,]]} \\[2pt]
\textbf{Rejected} & \wt{58}{```}\wt{10}{json}\newline \wt{12}{[}\newline \wt{5}{\ }\wt{9}{\ ["}\wt{62}{ASS}\wt{10}{1}\wt{8}{",}\wt{9}{\ "}\wt{17}{si}\wt{10}{RNA}\wt{9}{",}\wt{9}{\ "}\wt{57}{NEG}\wt{6}{"],}\newline \wt{10}{\ }\wt{10}{\ ["}\wt{12}{b}\wt{10}{ort}\wt{10}{ez}\wt{10}{om}\wt{10}{ib}\wt{10}{",}\wt{9}{\ "}\wt{5}{cis}\wt{10}{pl}\wt{10}{atin}\wt{12}{",}\wt{10}{\ "}\wt{0}{p}\wt{10}{emet}\wt{10}{rex}\wt{10}{ed}\wt{10}{",}\wt{10}{\ "}\wt{40}{POS}\wt{9}{"]}\newline \wt{10}{]}\newline \wt{9}{```}\wt{10}{<|endoftext|>} \\[2pt]
\textbf{Errors} & The \texttt{```} fence breaks the instruction to return only the array; \texttt{ASS1} is a gene and \texttt{siRNA} a technique, so the first combination is fabricated; and the genuine combination is labelled \textsc{pos} rather than \textsc{comb}. \\[2pt]
\midrule
\multicolumn{2}{@{}l@{}}{\textbf{(b) Structured numerical answer}} \\[2pt]
\textbf{Prompt} & Flights of 8, 4 and 6 hours with a 2-hour delay at each border. Report the total travel time as JSON. \\[2pt]
\textbf{Chosen} & \texttt{```json\newline\{"total\_travel\_time\_hours": 24\}\newline```} \\[2pt]
\textbf{Rejected} & \wt{11}{```}\wt{12}{json}\newline \wt{11}{\{}\newline \wt{11}{\ }\wt{12}{\ "}\wt{11}{total}\wt{11}{\_time}\wt{11}{":}\wt{62}{\ "}\wt{60}{18}\wt{11}{\ hours}\wt{50}{",}\newline \wt{11}{\ }\wt{11}{\ "}\wt{34}{border}\wt{11}{\_del}\wt{12}{ays}\wt{0}{":}\wt{24}{\ }\wt{40}{2}\newline \wt{11}{\}}\newline \wt{11}{```}\wt{11}{<|endoftext|>} \\[2pt]
\textbf{Errors} & The total is returned as a string rather than a number, \texttt{18} counts only the flights and drops the delays, and \texttt{border\_delays} is a field nobody asked for. \\[2pt]
\midrule
\multicolumn{2}{@{}l@{}}{\textbf{(c) Chemical classification}} \\[2pt]
\textbf{Prompt} & Classify each oxide as strongly acidic, weakly acidic, neutral, amphoteric, weakly basic or strongly basic: \dots\ (j)~$\mathrm{OsO_4}$ (k)~$\mathrm{Al_2O_3}$ \dots \\[2pt]
\textbf{Chosen} & \textbf{(j)} $\mathrm{OsO_4}$ -- strongly \emph{acidic}; forms osmic acid in water. \quad \textbf{(k)} $\mathrm{Al_2O_3}$ -- \emph{amphoteric}; reacts with both acid and base. \\[2pt]
\textbf{Rejected} & \textcolor{gray}{[\dots]}\ \wt{8}{(j}\wt{8}{)}\wt{8}{\ Os}\wt{8}{O}\wt{7}{\_}\wt{8}{4}\wt{8}{\ (}\wt{19}{Os}\wt{9}{O}\wt{8}{):}\wt{5}{\ Strong}\wt{8}{ly}\wt{62}{\ basic}\wt{7}{\ (}\wt{6}{acid}\wt{7}{ic}\wt{7}{)}\newline \wt{8}{(k}\wt{8}{)}\wt{8}{\ Al}\wt{7}{\_}\wt{8}{2}\wt{8}{O}\wt{7}{\_}\wt{8}{3}\wt{8}{\ (}\wt{8}{Al}\wt{50}{uminum}\wt{7}{\ oxide}\wt{8}{):}\wt{42}{\ Strong}\wt{8}{ly}\wt{0}{\ basic}\wt{8}{\ (}\wt{46}{acid}\wt{7}{ic}\wt{8}{)}\newline \ \textcolor{gray}{[\dots]} \\[2pt]
\textbf{Errors} & Both oxides are called \emph{strongly basic}: $\mathrm{OsO_4}$ is acidic and $\mathrm{Al_2O_3}$ amphoteric. Each line then contradicts its own label in the following parenthesis. \\[2pt]
\bottomrule
\end{tabularx}
\end{table}

First, larger weights tend to appear near the point where the rejected response diverges from the preferred behavior, rather than on tokens that follow from that error. In (a), the opening fence receives a larger weight, while the closing fence will not be penalized as harshly. Similarly, \texttt{ASS} and \texttt{NEG}, which introduce and label a fabricated combination, receive high weights, whereas the tokens completing \texttt{ASS1} and \texttt{siRNA} receive low weights, as they are a natural completion after the starting tokens.

Second, correctly reproduced content generally receives small weights. The copied drug names in (a), the JSON structure in (a) and (b), and the formulas in (c) all receive low weights. Under standard DPO, these tokens would receive the same rejected-response penalty as the erroneous tokens.

Third, the weights depend on the token's role relative to the chosen response rather than on its surface form alone. In (c), \texttt{acid} receives a low weight where it is correct and high where it is incorrect. The same fence token receives a large weight in (a), where it violates an explicit instruction, and low in (b), where no such instruction is present.

\subsection{Uniform and Random Weight Controls}
\label{app:controls}

Because $w_t \leq 1$, our weighted objective reduces the total negative contribution of rejected tokens relative to standard DPO. To separate the effect of this overall attenuation from the effect of assigning different weights to different tokens, we train two controls at $\beta = 0.002$ using the setup of Section~\ref{sec:hyperparameters} (the same setup as our best model variant). \textbf{Uniform} assigns $w_t = 0.489$, the mean of our learned weights, to every rejected token. \textbf{Random} samples $w_t \sim \mathcal{U}[0,1]$ independently for each token once before training, matching the expected attenuation while removing any relationship between the weight and the token. Table~\ref{tab:weight_ablation} reports the results.

\begin{table}[t]
    \centering
    \small
    \setlength{\tabcolsep}{5pt}
    \caption{
        Comparison of gradient-based, uniform, and random rejected-token weights.
        We report mean performance (\%) across 3 runs, with standard deviation shown in gray.
        Best results per benchmark are \textbf{bolded}.
    }
    \label{tab:weight_ablation}
    \begin{tabular}{lccc}
        \toprule
        \textbf{Benchmark}
        & \textbf{GAW-PO}
        & \textbf{Uniform Weights}
        & \textbf{Random Weights} \\
        \midrule

        AIME 2024
        & 16.77 \std{0.79}
        & \textbf{17.12} \std{0.79}
        & 16.74 \std{0.52} \\

        AIME 2025
        & \textbf{17.53} \std{0.68}
        & 16.91 \std{0.37}
        & 17.01 \std{0.81} \\

        MATH
        & 76.63 \std{0.39}
        & \textbf{76.71} \std{0.35}
        & 76.61 \std{0.36} \\

        Omega 500
        & 12.33 \std{0.70}
        & 13.20 \std{0.92}
        & \textbf{13.53} \std{0.81} \\

        \midrule

        AGI Eval
        & \textbf{65.49} \std{0.24}
        & 65.12 \std{0.41}
        & 64.26 \std{0.67} \\

        BBH (CoT)
        & \textbf{67.79} \std{0.49}
        & 64.96 \std{0.40}
        & 63.96 \std{0.14} \\

        \midrule

        HumanEval+
        & \textbf{75.35} \std{0.57}
        & 62.40 \std{0.88}
        & 53.88 \std{0.36} \\

        LiveCodeBench
        & \textbf{30.01} \std{0.79}
        & 28.01 \std{0.19}
        & 26.15 \std{0.25} \\

        MBPP+
        & \textbf{59.11} \std{0.29}
        & 55.85 \std{0.56}
        & 54.20 \std{0.21} \\

        \midrule

        GPQA
        & \textbf{35.04} \std{0.39}
        & 34.75 \std{1.52}
        & 34.82 \std{1.24} \\

        MMLU (CoT)
        & \textbf{69.98} \std{0.57}
        & 68.03 \std{0.35}
        & 67.86 \std{0.28} \\

        \midrule
        \textbf{Average}
        & \textbf{47.82}
        & 45.73
        & 44.46 \\

        \bottomrule
    \end{tabular}
\end{table}

Uniform weighting recovers much of the stability, suggesting that reducing the magnitude of the negative update is itself beneficial. However, attenuation alone does not explain the full improvement: gradient-aligned weights outperform uniform weights by 2.09 points and random weights by 3.36 points.

The largest differences appear on tasks where rejected responses can contain substantial correct structure shared with the chosen response. Relative to GAW-PO, uniform and random weighting are lower by 12.95/21.47 points on HumanEval+, 2.83/3.83 points on BBH, and 1.95/2.12 points on MMLU, while the mathematics benchmarks are generally within one standard deviation across variants. Random weighting also performs below uniform weighting overall. These controls suggest that performance at small $\beta$ certainly depends on assigning the penalty selectively across tokens.

\section{Additional Experiments - Detailed Results}
\label{app:additional_experiments}
\subsection{Category Breakdown of Robustness to the DPO Regularization Strength}
\label{app:beta_sweep_results}

\begin{figure*}[t]
    \centering
    \includegraphics[width=\textwidth]{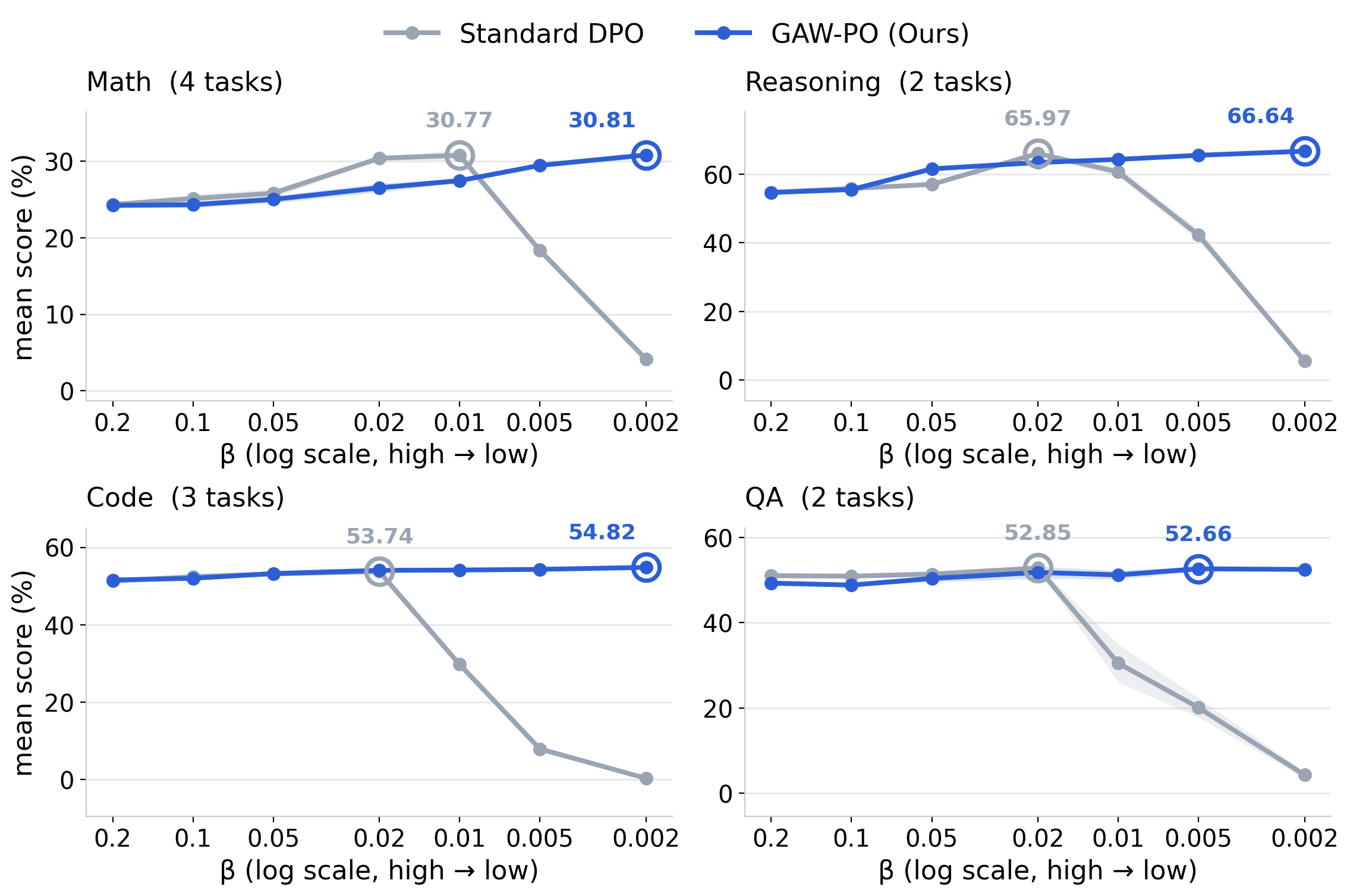}
    \caption{
    Performance across $\beta$ grouped by benchmark category. Standard DPO degrades sharply under aggressive preference optimization across all categories, while our method remains stable and generally improves as $\beta$ decreases.
    }
    \label{fig:tasks}
\end{figure*}

This section provides a more detailed view of the $\beta$ sweep presented in Section~\ref{sec:beta_robustness}. Figure~\ref{fig:tasks} breaks down the aggregate results by benchmark category, showing how standard DPO and GAW-PO behave across mathematics, reasoning, coding, and question answering as the optimization becomes more aggressive.

Standard DPO is competitive, and occasionally better, around $\beta=0.02$, but its performance collapses across all four categories as $\beta$ becomes smaller. The effect is particularly severe for coding, reasoning, and QA, where the scores approach the bottom of their respective ranges at $\beta=0.002$. Our method, in contrast, remains stable across all categories. Mathematics, reasoning, and coding improve consistently toward smaller values of $\beta$, while QA remains approximately stable despite the increasingly aggressive updates.

\subsection{Full Cluster-Variant Results}
\label{app:cluster_results}

Table~\ref{tab:cluster_full_results} reports the complete benchmark results corresponding to the aggregate scores shown in Table~\ref{tab:cluster_results} of the main paper.

\begin{table}[t]
\centering
\small
\setlength{\tabcolsep}{5pt}
\caption{
Full benchmark results for the single dataset-level mean and clustered
preferred-direction variants, all trained at $\beta=0.002$. Values are
percentages, mean $\pm$ standard deviation over 3 seeds.
Bold marks the highest value in each row.
}
\label{tab:cluster_full_results}
\begin{tabular}{lcccc}
\toprule
\textbf{Benchmark}
& \textbf{Mean}
& \textbf{$K$=2}
& \textbf{$K$=4}
& \textbf{$K$=8}
\\
\midrule

AIME 2024
& 16.77 {\scriptsize $\pm$ 0.79}
& \textbf{17.67} {\scriptsize $\pm$ 0.51}
& 17.29 {\scriptsize $\pm$ 1.08}
& 16.01 {\scriptsize $\pm$ 0.61}
\\

AIME 2025
& 17.53 {\scriptsize $\pm$ 0.68}
& 17.36 {\scriptsize $\pm$ 0.57}
& \textbf{17.81} {\scriptsize $\pm$ 0.28}
& 17.08 {\scriptsize $\pm$ 0.28}
\\

MATH
& 76.63 {\scriptsize $\pm$ 0.39}
& 76.81 {\scriptsize $\pm$ 0.40}
& \textbf{76.98} {\scriptsize $\pm$ 0.17}
& 76.48 {\scriptsize $\pm$ 0.79}
\\

Omega 500
& 12.33 {\scriptsize $\pm$ 0.70}
& \textbf{14.20} {\scriptsize $\pm$ 0.53}
& 12.27 {\scriptsize $\pm$ 0.64}
& 12.93 {\scriptsize $\pm$ 0.64}
\\
\midrule

AGI Eval
& 65.49 {\scriptsize $\pm$ 0.24}
& \textbf{66.37} {\scriptsize $\pm$ 0.71}
& 65.52 {\scriptsize $\pm$ 0.28}
& 65.71 {\scriptsize $\pm$ 0.48}
\\

BBH (CoT)
& \textbf{67.79} {\scriptsize $\pm$ 0.49}
& 67.24 {\scriptsize $\pm$ 0.93}
& 67.26 {\scriptsize $\pm$ 0.59}
& 67.57 {\scriptsize $\pm$ 0.77}
\\
\midrule

HumanEval+
& 75.35 {\scriptsize $\pm$ 0.57}
& 63.01 {\scriptsize $\pm$ 1.40}
& \textbf{75.63} {\scriptsize $\pm$ 0.79}
& 75.61 {\scriptsize $\pm$ 0.88}
\\

LiveCodeBench
& 30.01 {\scriptsize $\pm$ 0.79}
& 28.21 {\scriptsize $\pm$ 0.69}
& 30.06 {\scriptsize $\pm$ 0.21}
& \textbf{30.17} {\scriptsize $\pm$ 0.56}
\\

MBPP+
& \textbf{59.11} {\scriptsize $\pm$ 0.29}
& 57.05 {\scriptsize $\pm$ 0.27}
& 58.84 {\scriptsize $\pm$ 0.29}
& 59.07 {\scriptsize $\pm$ 0.74}
\\
\midrule

GPQA
& 35.04 {\scriptsize $\pm$ 0.39}
& 34.90 {\scriptsize $\pm$ 1.29}
& 35.94 {\scriptsize $\pm$ 2.32}
& \textbf{36.38} {\scriptsize $\pm$ 1.12}
\\

MMLU (CoT)
& \textbf{69.98} {\scriptsize $\pm$ 0.57}
& 69.79 {\scriptsize $\pm$ 0.27}
& 69.74 {\scriptsize $\pm$ 0.38}
& 69.82 {\scriptsize $\pm$ 0.17}
\\
\midrule

\textbf{Average}
& 47.82 
& 46.60 
& \textbf{47.94} 
& 47.89 
\\
\bottomrule
\end{tabular}
\end{table}

The results reinforce the conclusion that additional preferred directions provide only modest gains over the single dataset-level mean. $K=4$ obtains the strongest aggregate score, while $K=8$ is close behind and $K=2$ performs worse. No value of $K$ dominates across all benchmarks.

To understand this, we further measure which component attains the maximum alignment for each rejected token. In the original formulation, the paired chosen direction dominates for $79.16\%$ of tokens, while the dataset mean dominates for $20.84\%$. With $K=2$, $K=4$, and $K=8$, the paired direction dominates for $66.69\%$, $62.52\%$, and $53.18\%$ of tokens, respectively, meaning that cluster-level directions determine the score for up to $46.82\%$ of rejected tokens.

This shows that the clustered directions are actively used and capture preferred update directions not represented by the paired response alone. However, greater cluster utilization does not directly imply better downstream performance: $K=8$ uses clustered directions most frequently but performs slightly below $K=4$, while $K=2$ underperforms the original formulation. Simply increasing the number of reference directions therefore does not monotonically improve the quality of the token weights; the quality of the preferred-direction decomposition matters more than the number of available directions.

\section{Limitations}
\label{app:limitations}

Our method adds an offline stage that computes rejected-token weights before training. This stage is performed once per dataset and probe and is reused across all training runs; in our experiments it costs roughly 1.5 training runs. The cost is incurred once and is amortized across all subsequent training runs, hyperparameter sweeps, and ablations, as in our experiments. However, it is less lightweight than standard DPO for a single run.

Gradient alignment is measured in the parameter space of a low-rank PiSSA probe, which we use as a tractable proxy for the full parameter space. The choice is motivated by the relation between PiSSA gradients and full-parameter gradients, and the downstream results indicate that the resulting weights are informative. A systematic study of how the probe rank affects the weights, and of how closely low-rank alignment tracks full-parameter alignment, is left for future work.

The weights are computed from the SFT initialization and kept fixed during training. This matches the offline nature of DPO, where the preference pairs do not change, but the policy's gradient geometry does evolve over training. Estimating the weights dynamically is an extension and is required for the online setting discussed in Section~\ref{app:conclusion}.

The conversion of alignment scores into weights has more possible choices. Alternative mappings are a promising direction, including temperature or offset parameters in the sigmoid, normalized reference directions, and rank- or quantile-based mappings. 

Finally, our evaluation focuses on a single open model family and preference dataset, chosen to allow a controlled study, and on verifiable benchmarks in mathematics, reasoning, code, and question answering. We did not evaluate on judge-based open-ended benchmarks or on multi-turn and multilingual preference data, and extending the analysis to these settings and to larger models is an important direction for future work.

\end{document}